\documentclass[letterpaper]{article} 
\usepackage{aaai2026}  
\usepackage{times}  
\usepackage{helvet}  
\usepackage{courier}  
\usepackage[hyphens]{url}  
\usepackage{graphicx} 
\usepackage{natbib}  
\usepackage{caption} 
\usepackage{float}
\usepackage{newfloat}
\usepackage{listings}
\DeclareCaptionStyle{ruled}{labelfont=normalfont,labelsep=colon,strut=off} 
\floatstyle{ruled}
\newfloat{listing}{tb}{lst}{}
\floatname{listing}{Listing}

\usepackage{multirow} 
\usepackage{subfig}
\usepackage{booktabs}
\usepackage{amsmath}
\usepackage{amsfonts}
\usepackage{colortbl}
\usepackage[table,xcdraw]{xcolor}
\definecolor{mygray}{gray}{.875}

\definecolor{lightblue}{RGB}{239,245,251}  
\usepackage{pifont}
\usepackage[T1]{fontenc}
\usepackage{pifont}
\usepackage{fontawesome}

\title{Decoupling What to Count and Where to See for Referring Expression Counting}
\author {
        Yuda Zou,
    Zijian Zhang,
    Yongchao Xu\textsuperscript{\faEnvelope}\\
}
\affiliations{
    School of Computer Science, Wuhan University, Wuhan, China\\
    zouyuda@whu.edu.cn
}

\begin{document}

\maketitle

\begin{abstract}
Referring Expression Counting (REC) extends class-level object counting to the fine-grained subclass-level, aiming to enumerate objects matching a textual expression that specifies both the class and distinguishing attribute.
A fundamental challenge, however, has been overlooked:
annotation points are typically placed on class-representative locations (\textit{e.g.}, heads), forcing models to focus on class-level features while neglecting attribute information from other visual regions (\textit{e.g.}, legs for ``walking'').
To address this, we propose \textit{W2-Net}, a novel framework that explicitly decouples the problem into ``what to count'' and ``where to see'' via a dual-query mechanism.
Specifically, alongside the standard what-to-count (w2c) queries that localize the object, we introduce dedicated where-to-see (w2s) queries. 
The w2s queries are guided to seek and extract features from attribute-specific visual regions, enabling precise subclass discrimination.
Furthermore, we introduce Subclass Separable Matching (SSM), a novel matching strategy that incorporates a repulsive force to enhance inter-subclass separability during label assignment. 
\textit{W2-Net} significantly outperforms the state-of-the-art on the REC-8K dataset, reducing counting error by 22.5\%~(validation) and 18.0\%~(test), and improving localization F1 by 7\% and 8\%, respectively. 
Code will be available.
\end{abstract}

\section{Introduction}
\label{sec: introduction}

\begin{figure}[!t]
  \centering
  \includegraphics[width=1.\linewidth]{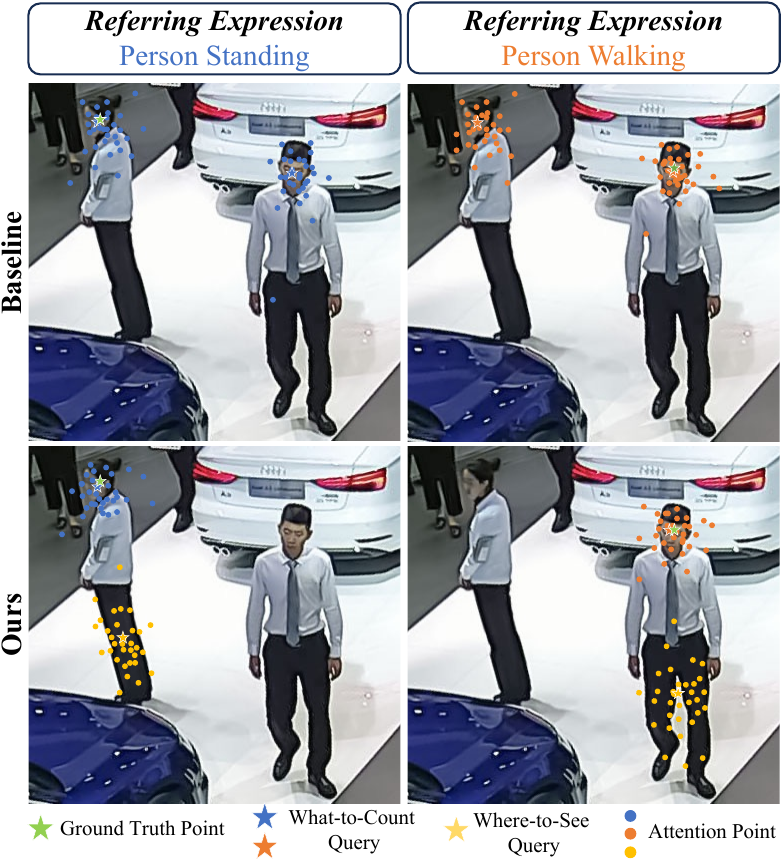}
\caption{\textbf{Illustration of the core challenge in REC we address.} 
REC point annotations (green pentastar), placed on class-representative locations like heads, provide insufficient guidance for attribute-specific regions (\textit{e.g.}, legs for ``standing'' or ``walking''). 
This hinders the model from distinguishing fine-grained subclasses (``person standing'' and ``person walking'').
Our \textit{W2-Net} introduces a dedicated where-to-see (w2s) queries (yellow pentastar) that actively seek attribute-relevant visual cues.
By fusing these features to the corresponding standard what-to-count (w2c), the model achieves precise subclass discrimination. 
The attention points visualize the attention focus of each query type. Best viewed by zooming in the electronic version.
}
\label{fig: motivation}
\end{figure}

Object counting, a fundamental computer vision task, has diverse applications from crowd analysis~\cite{han2023steerer_STEERER, huang2024count_action} and traffic management~\cite{urban_planning} to cell biology~\cite{chen2021cell_yajie}. 
Early work in this domain, known as Class-Specific Counting (CSC), developed specialized models for predefined classes. 
Among these, Crowd Counting~\cite{song2021rethinking_P2P} has been the most extensively studied, aiming to determine the number of people in an image.
Crowd counting methods can generally be divided into density-based ones~\cite{lin2024gramformer, yang2025taste} and localization-based ones~\cite{song2021rethinking_P2P, lin2025point, liang2022end_CLTR}. 
While effective, CSC methods are limited in flexibility, requiring a new model to be trained for each new class of interest. 
This spurred the development of Class-Agnostic Counting (CAC)~\cite{ranjan2021learning_FSC}, which generalizes to arbitrary classes using visual~\cite{yang2025pbecount} or text prompts~\cite{shi2025text, wang2024enhancing}.

Despite these progress, both CSC and CAC models are limited to count objects at the class level, remaining oblivious to the fine-grained attributes that define distinct subclasses within a class
to bridge this gap. Referring Expression Counting (REC)~\cite{dai2024referring_REC} has recently emerged, aiming to count objects of a specific subclass defined by a given textual referring expression (\textit{e.g.}, ``people walking'' vs. ``people standing''), which specifies both a class name and a distinguishing attribute.
The key challenge in REC lies in distinguishing different subclasses that share the same class but differ in specific attributes~\cite{triaridis2025improving_REC_newwork}. 
To alleviate this, the pioneering work, GroundingREC~\cite{dai2024referring_REC}, adopts a detection-then-count pipeline built on the powerful open-set detector GroundingDINO~\cite{liu2024grounding_gdino}.
It enhances the detector with global-local feature fusion and contrastive learning to differentiate the attributes.
Subsequently, CAD-GD~\cite{wang2025exploring_CAD-GD} proposes a density-based module to guide the localization and counting process.
While they have made promising strides, they overlook a critical limitation inherent in the REC's annotation scheme.

Following standard practice in object counting datasets~\cite{zhang2016single_SH_MCNN, idrees2018composition_UCF-QNRF, sindagi2020jhu}, REC dataset~\cite{dai2024referring_REC} uses a cost-effective point annotation format where each target object is marked with a single 2D point. 
These points are typically marked on class-representative locations like a person's head (see Fig.~\ref{fig: motivation})
without considering the visual regions that characterize the specific attributes. 
Used for supervision, these annotation points inherently force the model to over-emphasize class-level features extracted from their vicinity, while neglecting the crucial attribute-specific features located elsewhere (\textit{e.g.}, legs for ``walking''). 
As illustrated in Fig.~\ref{fig: motivation}, this lack of attribute-aware guidance makes it difficult for the model to distinguish between different subclasses (\textit{e.g.}, ``person walking'' and ``person standing'') and ultimately impairs the counting performance.

To explicitly tackle this challenge, we propose a novel framework \textit{W2-Net}. 
Following prior methods~\cite{dai2024referring_REC, wang2025exploring_CAD-GD}, \textit{W2-Net} is built upon the powerful GroundingDINO detector~\cite{liu2024grounding_gdino}, adapting it from box detection to point-based localization for the counting task. 
At the core of \textit{W2-Net} is a novel W2 decoder, designed to decouple the problem into ``what to count'' and ``where to see''.
Alongside the standard what-to-count (w2c) query that aims to localize the target object, we additionally introduce a dedicated where-to-see (w2s) query. Crucially, both queries are processed in parallel and iteratively refined through each decoder layer. 
While the w2c query learns to converge on the class-representative center (\textit{e.g.}, a person's head) consistent with the supervised annotation point, the w2s query is specifically guided by the attribute portion in the referring expression. 
This guidance enables it to actively navigate the feature space and anchor itself to the visual region most indicative of the given attribute (\textit{e.g.}, yellow pentastar and attention points on legs for "walking" and ``standing'' in Fig.~\ref{fig: motivation}). 
The features from these regions are then extracted by the w2s query and fused to enrich the w2c query, enabling improved subclass distinction and counting.
Furthermore, we introduce Subclass Separable Matching (SSM) to alleviate the training instability arising from inter-subclass ambiguity. By incorporating a repulsive force into the matching cost, SSM actively penalizes assignments of w2c queries near non-target subclasses, thereby enforcing greater separability and ensuring more stable supervision.

Extensive experiments on the REC-8K benchmark~\cite{dai2024referring_REC} demonstrate that our method significantly outperforms the state-of-the-art methods on both counting and localization.
Specifically, our approach achieves a remarkable reduction in relative counting error by $22.5\%$ on the validation set and $18.0\%$ on the test set. Concurrently, it boosts the localization F1-score by $7\%$ and $8\%$, respectively. 
Our contributions are three-fold:
\begin{itemize}
\item To the best of our knowledge, we are the first to identify and analyze the annotation issue in REC.
\item We propose a novel framework, \textit{W2-Net}, and a subclass separable matching, effectively enhancing the attribute perception and subclass discrimination.
\item We set a new state-of-the-art on REC, significantly outperforming prior methods in counting and localization.
\end{itemize}

\section{Related Work}
\label{sec: related work}

\subsection{The Evolution of Object Counting}
\label{subsec: related work the Evolution of Object Counting}

\begin{figure*}[!t]
  \centering
  \includegraphics[width=1.\linewidth]{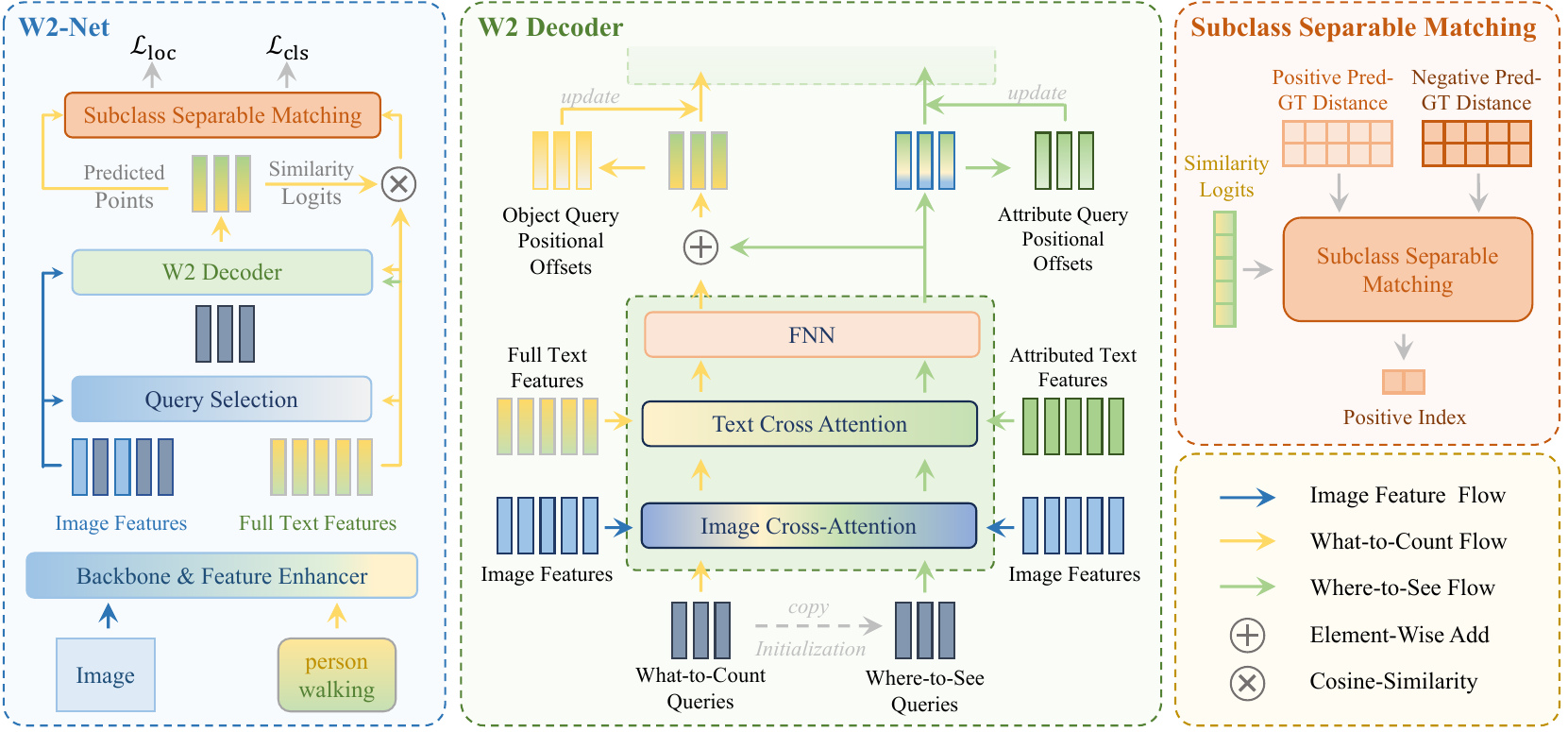}
  \caption{\textbf{The framework of \textit{W2-Net}.} \textit{W2-Net} decouples ``what to count'' and ``where to see'' in the proposed W2 Decoder, where the \textbf{what-to-count (w2c) query} targets at locating the object's class-representative center and the parallel dedicated \textbf{where-to-see (w2s) query} grounds the distinguishing attribute. Fusing their features enables precise subclass discrimination. Besides, we develop the \textbf{Subclass Separable Matching (SSM)} to stabilize training by introducing a repulsive force into the matching cost, effectively resolving inter-subclass ambiguity and ensuring stable supervision.
  }
  \label{fig: pipeline}
\end{figure*}

The field of object counting~\cite{pothiraj2025capture, dumery2024counting} has evolved significantly, moving from constrained, class-specific counting to more versatile and flexible paradigms~\cite{mondal2025omnicount, guo2025enhancing_pruning, perez2024discount_buildingcount, he2024soundcount}.
Initial efforts focused on \textbf{Class-Specific Counting (CSC)}, where models were tailored to enumerate instances of a single, pre-defined class, such as people~\cite{han2023steerer_STEERER} or vehicles~\cite{hsieh2017drone_carpk}. 
These methods, typically based on indirect density map regression~\cite{sun2023indiscernible_underwater} or direct object localization~\cite{song2021rethinking_P2P, liu2023point_PET}, achieved impressive performance.
However, they are inherently constrained in the fixed target classes, lacking the ability to generalize to more.
This fundamental limitation spurred the development of \textbf{Class-Agnostic Counting (CAC)}~\cite{ranjan2021learning_FSC}, which enables counting objects of arbitrary classes at test time prompted by visual exemplars~\cite{chen2025single} or a class name~\cite{lin2024fixed, qian2025t2icount}.
Despite this significant leap in flexibility for counting, CAC models still operate at the class level, remaining oblivious to fine-grained attributes that distinguish different subclasses within a class.
To bridge this gap, \textbf{Referring Expression Counting (REC)}~\cite{dai2024referring_REC} was recently introduced to tackle the more challenging problem of subclass counting.
REC enables the enumeration of specific object subclasses defined by textual referring expressions, marking a crucial progression towards more intelligent and practical counting systems.

\subsection{Referring Expression Counting}
\label{subsec: related work referring expression counting}
Referring Expression Counting (REC) moves beyond class-level counting to the more challenging subclass level. 
The goal is to enumerate objects that match a given textual description, which specifies both a class name and the distinguishing attribute.
The core challenge of REC, therefore, is to accurately distinguish between subclasses that share the same parent class but differ in specific attributes.
The pioneering work, GroundingREC~\cite{dai2024referring_REC}, introduced a detection-then-count pipeline based on the open-set detector GroundingDINO~\cite{liu2024grounding_gdino}.
It enhances the detector with global-local feature fusion and contrastive learning to improve attribute differentiation. 
Following this, CAD-GD~\cite{wang2025exploring_CAD-GD} proposed a contextual attribute density module to guide localization.
While these methods have made promising strides, they inherit and overlook a critical issue rooted in the standard annotation protocol of the REC dataset~\cite{dai2024referring_REC}.

\subsection{The Challenge of Point Annotations in Counting}
\label{subsec: The Challenge of Point Annotations in Counting}
Due to their cost-effectiveness, 2D point annotations are the predominant format in object counting. 
In conventional class-level counting, these points are intended to mark class-representative locations (\textit{e.g.}, persons' heads). 
However, they are often prone to positional errors arising from annotator subjectivity.
Several methods have been proposed to mitigate this issue. 
For density-based counting, robust loss functions like Bayesian Loss~\cite{ma2019bayesian_BL} and NoiseCC~\cite{wan2023modeling_NoiseCC_Tpami} were designed to handle noisy density maps from noisy annotation points. 
For localization-based counting, where these density-based solutions are less applicable, SAE~\cite{zou2024_SAE} was proposed to directly refine the point annotations by alleviating their positional inconsistency. 
Crucially, these methods primarily address issues related to \textit{random positional noise}.

In REC, however, the annotation challenge transcends random noise.
While REC annotations also mark class-representative locations, the supervisory signal they provide is often spatially detached from the visual regions that characterize the subclass attribute (e.g., the legs for ``walking'' vs the head for ``person'').
This misalignment forces the model to choose between focusing on the class-representative location or seeking attribute-specific regions for correct subclass distinction, creating an inherent learning conflict.
Our work is the first to identify and explicitly tackle this contradiction, enabling the model to learn class features and discriminative subclass features simultaneously and effectively.

\section{Methodology}
\label{sec: methodology}

\subsection{Overview}
\label{subsec: overview}

This section details the proposed \textit{W2-Net}, a novel framework designed to tackle a fundamental yet overlooked challenge in REC: misalignment between class-representative annotation points and attribute-defining visual regions.
To address this, we introduce two synergistic modules: a novel W2 Decoder that decouples ``what to count'' and ``where to see'', and a Subclass Separable Matching (SSM) to ensure stable and accurate supervision.

Following prior methods~\cite{dai2024referring_REC, wang2025exploring_CAD-GD}, \textit{W2-Net} is built upon the open-set detector, GroundingDINO~\cite{liu2024grounding_gdino}, adapting it from box detection to point-based localization for the counting task. 
The overall architecture is depicted in Fig.~\ref{fig: pipeline}.
Given an input image $I$ and a referring expression text $T$ that specifies the class name and the attribute, we first extract image features $F_v \in \mathbb{R}^{M \times C}$ and text features $F_t \in \mathbb{R}^{N \times C}$ using a frozen backbone and a feature enhancer, where $M$ is the number of image tokens, $N$ is the number of text tokens, and $C$ is the feature dimension.
These features are then fed into our W2 Decoder. 
Here, dedicated \textit{where-to-see (w2s)} queries operate in parallel with the standard \textit{what-to-count (w2c)} queries.
The w2s queries are guided to locate attribute-relevant features, which are then fused to w2c queries to enable precise subclass discrimination.
Through the iterative refinement of multiple decoder layers, the enhanced w2c queries are passed to prediction heads to generate point locations and classification scores. 
During training, these predicted points are assigned to ground-truth labels via our Subclass Separable Matching (SSM), which introduces a repulsive force to mitigate matching ambiguity between similar subclasses.
Finally, the complete model is optimized using classification and localization losses.

\subsection{W2 Decoder}
\label{subsec: W2 Decoder}
As previously discussed, the central challenge in REC stems from the spatial misalignment between class-representative annotation points and attribute-representative visual regions. 
To address this, we propose W2 Decoder, which is explicitly designed to decouple the ``what to count'' and ``where to see''. 
It introduces a dedicated where-to-see (w2s) query ($Q_{w2s}$) that operates in parallel with the standard what-to-count (w2c) query ($Q_{w2c}$).
While the w2c query learns to converge on the annotated point (``what to count''), the w2s query is specifically guided to seek out visual evidence for the given attribute (``where to see''). 
This parallel, specialized processing allows the model to learn both class-level and attribute-specific features effectively.

\smallskip
\noindent
\textbf{Query Initialization.} 
At the decoder's input (before the first decoder layer, $l=0$), we initialize a set of $K$ what-to-count (w2c) queries, where $K = 900$ as in GroundingDINO~\cite{liu2024grounding_gdino}. 
Each w2c query consists of a content embedding and a reference point.
Their content embeddings $\{q_{w2c, i}^0\}_{i=1}^K$ are learnable parameters, while their initial reference points $\{p_{w2c, i}^0\}_{i=1}^K$ are set to the spatial locations of the top-$K$ image features with the highest cross-modal similarity to the full text feature $F_t$.
We then initialize the where-to-see (w2s) queries by duplicating the w2c queries. 
Specifically, each w2s query's content embedding is a separate learnable parameter initialized from its w2c counterpart $q_{w2s, i}^0 = q_{w2c, i}^0$, and its reference point is directly copied $p_{w2s, i}^0 = p_{w2c, i}^0$.
This co-location initialization strategy ensures that the w2s query begins its search for attribute-specific visual cues from a sensible, class-relevant starting location.

\begin{table*}[t]
  \centering
\resizebox{.9\linewidth}{!}{
\begin{tabular}{l|c| c ||c c c c c}
\specialrule{0.2em}{0pt}{0pt}

\rowcolor{mygray}      &      &       & \multicolumn{5}{c}{Validation Set}        \\ 
\cline{4-8}
\rowcolor{mygray}       \multirow{-2}{*}{Method}           &    \multirow{-2}{*}{Venue}      &  \multirow{-2}{*}{Backbone}               & MAE~$\downarrow$            & RMSE~$\downarrow$           & Prec~$\uparrow$ & Rec~$\uparrow$  & F1~$\uparrow$     \\ 
\specialrule{0.15em}{0pt}{0pt}
ZSC~\cite{xu2023zero_ZSC}    &  CVPR23           & Swin-T                & 12.96         & 26.74          & -    & -    & -     \\
CounTX~\cite{AminiNaieni23_CounTX}     &   BMVC23      & ViT-B              & 11.88         & 27.04          & -    & -    & -     \\
GroundingDINO~\cite{liu2024grounding_gdino}   &  ECCV24  & Swin-T                & 9.03          & 21.98          & 0.56 & \underline{0.76} & 0.65  \\
GroundingREC~\cite{dai2024referring_REC}   & CVPR24    & Swin-T                & 6.80          & 18.13          & 0.65 & 0.71 & 0.68  \\
CAD-GD~\cite{wang2025exploring_CAD-GD}  & CVPR25    & Swin-T                & 5.43          & 15.01          & 0.68 & 0.72 & 0.70  \\
CAD-GD\dag{}~\cite{wang2025exploring_CAD-GD} &  CVPR25    & Swin-T                &\underline{4.58}         & \underline{13.24}          & \underline{0.68} & 0.71 & \underline{0.70}  \\ 

\rowcolor{lightblue} \textbf{W2-Net (Ours)}  &  -   & Swin-T                &   \textbf{3.55}      &   \textbf{10.39}     &  \textbf{0.76}& \textbf{0.78}   & \textbf{0.77}  \\ 

\specialrule{0.15em}{0pt}{0pt}

GroundingREC~\cite{dai2024referring_REC}   &  CVPR24  & Swin-B                & 5.66          & 15.24          & 0.66 & \underline{0.77} & 0.71  \\
CAD-GD~\cite{wang2025exploring_CAD-GD}  & CVPR25 & Swin-B                & 4.83          & 13.52          & 0.74 & 0.76 & \underline{0.75}  \\
CAD-GD\dag{}~\cite{wang2025exploring_CAD-GD} & CVPR25  & Swin-B                & \underline{4.23} & \underline{13.14} & \underline{0.76} & 0.70 & 0.73  \\ 
\rowcolor{lightblue} \textbf{W2-Net (Ours)}  & - & Swin-B        &   \textbf{3.47}     & \textbf{9.55}        &   \textbf{0.79}     & \textbf{0.79} & \textbf{0.79}   \\ 

\specialrule{0.2em}{0pt}{0pt}

\end{tabular}}
\caption{\textbf{Comparison with state-of-the-art approaches on the REC-8K validation set~\cite{dai2024referring_REC}.} The best and second best are \textbf{boldfaced} and \underline{underlined}, respectively. \dag{} indicates using a density-based query selection strategy.}

\label{tab: rec-8k val set}
\end{table*}

\begin{figure}[!t]
  \centering
  \includegraphics[width=1.\linewidth]{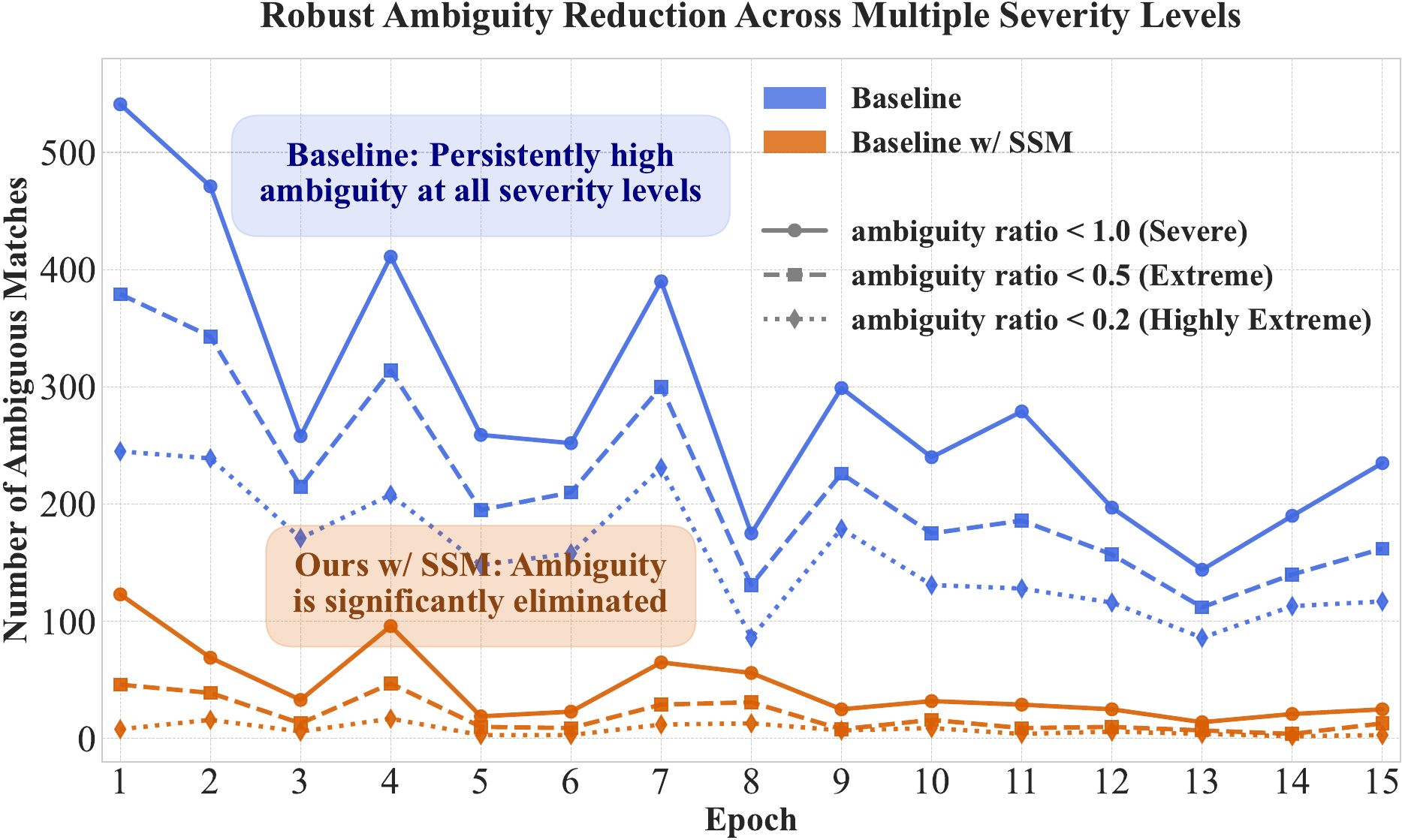}
\caption{\textbf{Effectiveness of Subclass Separable Matching (SSM) in resolving training ambiguity.} 
A standard matching approach (blue) consistently suffers from high ambiguity due to inter-subclass similarity. In contrast, our SSM (orange), which incorporates a repulsive force, alleviates such ambiguity from the beginning of training, ensuring a stable and accurate supervision signal and improved performance.
}
\label{fig: subclass separable matching}
\end{figure}

\begin{table*}[t]
\centering
\resizebox{.9\linewidth}{!}{
\begin{tabular}{l|c| c ||c c c c c}
\specialrule{0.2em}{0pt}{0pt}

\rowcolor{mygray}      &      &       & \multicolumn{5}{c}{Test Set}        \\ 
\cline{4-8}
\rowcolor{mygray}       \multirow{-2}{*}{Method}           &    \multirow{-2}{*}{Venue}      &  \multirow{-2}{*}{Backbone}               & MAE~$\downarrow$   & RMSE~$\downarrow$  & Prec~$\uparrow$ & Rec~$\uparrow$  & F1$\uparrow$     \\ 
\specialrule{0.15em}{0pt}{0pt}
ZSC~\cite{xu2023zero_ZSC}    &  CVPR23           & Swin-T                & 13.00 & 29.07 & -    & -    & -     \\
CounTX~\cite{AminiNaieni23_CounTX}     &   BMVC23      & ViT-B              & 11.84 & 25.62 & -    & -    & -     \\
GroundingDINO~\cite{liu2024grounding_gdino}   &  ECCV24  & Swin-T                & 8.88  & 21.95 & 0.59 & \underline{0.76} & 0.66  \\
GroundingREC~\cite{dai2024referring_REC}   & CVPR24    & Swin-T                & 6.50  & 19.79 & 0.67 & 0.72 & 0.69  \\
CAD-GD~\cite{wang2025exploring_CAD-GD}  & CVPR25    & Swin-T                & 5.29  & 17.08 & 0.71 & 0.73 & \underline{0.72}  \\
CAD-GD\dag{}~\cite{wang2025exploring_CAD-GD} &  CVPR25    & Swin-T                & \underline{4.59}  & \underline{14.68} & \underline{0.72} & 0.70 & 0.71  \\ 

\rowcolor{lightblue} \textbf{W2-Net (Ours)}  &  -   & Swin-T                & \textbf{3.59}& \textbf{9.59}& \textbf{0.77} & \textbf{0.80} & \textbf{0.79}  \\ 

\specialrule{0.15em}{0pt}{0pt}

GroundingREC~\cite{dai2024referring_REC}   &  CVPR24  & Swin-B                & 5.42  & 18.47 & 0.71 & 0.69 & 0.70  \\
CAD-GD~\cite{wang2025exploring_CAD-GD}  & CVPR25 & Swin-B                & 4.94  & 14.65 & 0.75 & \underline{0.77} & \underline{0.76}  \\
CAD-GD\dag{}~\cite{wang2025exploring_CAD-GD} & CVPR25  & Swin-B                & \underline{4.34}  & \underline{12.93} & \underline{0.77} & 0.71 & 0.74  \\ 
\rowcolor{lightblue} \textbf{W2-Net (Ours)}  & - & Swin-B        & \textbf{3.56} &  \textbf{9.15} & \textbf{0.79} & \textbf{0.81} & \textbf{0.80}   \\ 

\specialrule{0.2em}{0pt}{0pt}

\end{tabular}}
\caption{\textbf{Comparison with state-of-the-art approaches on the REC-8K test set.~\cite{dai2024referring_REC}} The best and second best are \textbf{boldfaced} and \underline{underlined}, respectively. \dag{} indicates using a density-based query selection strategy.}
\label{tab: rec-8k test set}
\end{table*}

\smallskip
\noindent
\textbf{Parallel Query Refinement.} 
Within each decoder layer $l$, the w2c queries and the w2s queries are refined in parallel with different focuses.

For the \textit{w2c query}, it first attends to the full text features $F_t$ to capture the complete semantic intent, and then to the image features $F_v$ via deformable attention to gather visual information around its current reference point $P_{w2c}$:
\begin{align}
    Q_{w2c}^{txt} &= \text{CrossAttn}(Q_{w2c}, F_t, F_t), \\
    Q_{w2c}^{img} &= \text{DeformAttn}(Q_{w2c}^{txt}, P_{w2c}, F_v).
\end{align}
The output is then processed by a Feed-Forward Network (FFN), yielding $\hat{Q}_{w2c} = \text{FFN}(Q_{w2c}^{img})$.

For the \textit{w2s query}, the process is similar but critically modified to isolate attribute-specific text features. It attends to a masked version of the text features, where class-related tokens are suppressed. Let $M_{w2s}$ be a binary mask that is one for attribute tokens and zero otherwise. This mask is straightforwardly derived from the provided annotations, which explicitly specify the class word for each referring expression, thus incurring no additional overhead.
The query is refined as:
\begin{align}
    Q_{w2s}^{txt} &= \text{CrossAttn}(Q_{w2s}, F_t \odot M_{w2s}, F_t \odot M_{w2s}), \\
    Q_{w2s}^{img} &= \text{DeformAttn}(Q_{w2s}^{txt}, P_{w2s}, F_v).
\end{align}
This forces the w2s query to focus exclusively on attribute-related words (\textit{e.g.}, ``walking''), guiding it to search for corresponding visual evidence specifically related to the attribute. 
This also yields an intermediate representation $\hat{Q}_{w2s} = \text{FFN}(Q_{w2s}^{img})$.

\smallskip
\noindent
\textbf{Fusion and Iterative Update.}
After parallel refinement, the attribute-centric information from the w2s query is injected into the w2c query stream to enhance its discriminative ability. This fusion is achieved via a simple element-wise addition. The fused representation is then passed through an FFN to produce the updated w2c query for the next layer ($l+1$):
\begin{equation}
    Q_{w2c}^{l+1} = \text{FFN}(\hat{Q}_{w2c}^{l} + \hat{Q}_{w2s}^{l}).
    \label{eq:fusion}
\end{equation}
Crucially, the w2s query is updated independently, preserving its specialized role:
\begin{equation}
    Q_{w2s}^{l+1} = \text{FFN}(\hat{Q}_{w2s}^{l}).
    \label{eq:attr_update}
\end{equation}
This independent update ensures that the w2s query remains a dedicated scout for attribute regions in the next layer, preventing its focus from being overly diluted by the w2c query's primary goal of localizing the object.

Finally, the reference points are updated. Two separate localization heads predict positional offsets from the refined queries, which are then added to the current reference points to obtain $P_{w2c}^{l+1}$ and $P_{w2s}^{l+1}$ for the next iteration. This iterative process allows w2c queries to converge on the annotated centers while w2s queries simultaneously navigate towards attribute-defining regions.

\begin{figure*}[t]
  \centering
  \includegraphics[width=1.\linewidth]{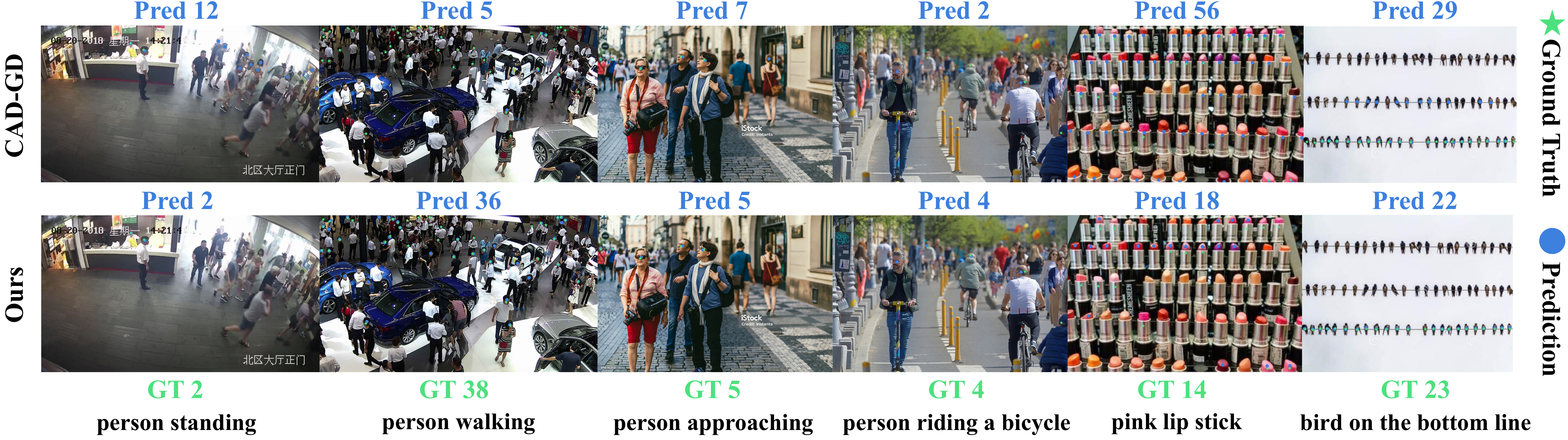}
  \caption{\textbf{Some qualitative results on the REC-8K dataset of CAD-GD~\cite{wang2025exploring_CAD-GD} and our W2-Net. }
}
\label{fig:qualitative_results}
\end{figure*}

\subsection{Subclass Separable Matching}
\label{subsec: subclass separable matching}
A standard one-to-one Hungarian matching~\cite{kuhn1955hungarian} based on classification and localization costs is suboptimal for REC. The high visual similarity between different subclasses (\textit{e.g.}, ``person walking'' vs. ``person standing'') often leads to matching ambiguity, where a query corresponding to an object of another subclass might be incorrectly matched to a ground truth, especially in early training stages (see Fig.~\ref{fig: subclass separable matching}). 
This results in unstable and noisy supervision.

To mitigate this, we propose Subclass Separable Matching (SSM), which introduces a repulsive force into the matching cost. 
This force explicitly penalizes a query assignment if the query is close to ground-truth points of other, non-target subclasses. Let $Y_{pos} = \{p_j\}_{j=1}^{N_{pos}}$ be the set of ground-truth points for the target subclass, and $Y_{neg} = \{p_k\}_{k=1}^{N_{neg}}$ be the ground-truth points for all other subclasses in the image. The matching cost $\mathcal{C}_{match}(i, j)$ between prediction $i$ and a ground-truth point $j \in Y_{pos}$ is:
\begin{equation}
    \mathcal{C}_{match}(i, j) = \lambda_{cls}(1 - \hat{s}_i) + \lambda_{L1}\|\hat{p}_i - p_j\|_1 + \lambda_{rep} \mathcal{C}_{rep}(\hat{p}_i, p_j),
    \label{eq:matching_cost}
\end{equation}
where $\hat{s}_i$ and $\hat{p}_i$ are the predicted score and location, and $\lambda$s are balancing weights (set to $\lambda_{cls}=5$ and $\lambda_{L1}=1$ as in prior works GroundingREC~\cite{dai2024referring_REC} and CAD-GD~\cite{wang2025exploring_CAD-GD}, and a hyperparameter $\lambda_{rep}$, respectively). 
The key component is our proposed repulsive cost $\mathcal{C}_{rep}$, which penalizes a query assignment if it is closer to a non-target subclass. To achieve this, we first define an ambiguity ratio, $\mathcal{R}(\hat{p}_i, p_j)$, which measures the relative proximity of a prediction $\hat{p}_i$ to the nearest non-target object versus its potential ground-truth match $p_j$:
\begin{equation}
    \mathcal{R}(\hat{p}_i, p_j) = \frac{\min_{p_k \in Y_{neg}} \|\hat{p}_i - p_k\|_2}{\|\hat{p}_i - p_j\|_2}~.
    \label{eq:ambiguity_ratio}
\end{equation}
The repulsive cost $\mathcal{C}_{rep}$ is then calculated based on this ratio:
\begin{equation}
    \mathcal{C}_{rep}(\hat{p}_i, p_j) = \exp(-\mathcal{R}(\hat{p}_i, p_j))~.
    \label{eq:repulsive_cost}
\end{equation}
The cost becomes large if a prediction $\hat{p}_i$ is closer to a negative object of another subclass (a distractor) than to its potential positive match $p_j$, thus discouraging ambiguous assignments. By incorporating this repulsive force, our matching enforces greater inter-subclass separability during the crucial label assignment phase, leading to more stable training and more discriminative final representations.
As illustrated in Fig.~\ref{fig: subclass separable matching}, our SSM effectively eliminates matching ambiguities in training, ensuring a more stable learning process.

\subsection{Training Objective}
\label{subsec: training objective}

Following the one-to-one label assignment by our Subclass Separable Matching, the network is optimized using a loss function that mirrors GroundingREC and CAD-GD. The total loss $\mathcal{L}$ is a weighted sum of a classification loss $\mathcal{L}_{cls}$ and a localization loss $\mathcal{L}_{loc}$:
\begin{equation}
    \mathcal{L} = \lambda_{cls} \mathcal{L}_{cls} + \lambda_{L1} \mathcal{L}_{loc},
    \label{eq:final_loss}
\end{equation}
where the weights $\lambda_{cls}$ and $\lambda_{loc}$ are consistent with those in the matching cost, set to $\lambda_{cls}=5$ and $\lambda_{L1}=1$ as in prior works. Specifically, we employ the focal loss for $\mathcal{L}_{cls}$ to handle the class imbalance between positive and negative queries, and the L1 loss for $\mathcal{L}_{loc}$ to point localization. 
Further details on the loss formulation are provided in the supplementary material.

\section{Experiments}
\label{sec: experimentation}

\subsection{Experimental Setup}
\label{subsec: Experimental Setup} 
\noindent
\textbf{Dataset and Metrics.} 
REC-8K~\cite{dai2024referring_REC}, the first REC benchmark, with 8,011 images and 17,122 image-RE pairs. Following prior work~\cite{dai2024referring_REC, wang2025exploring_CAD-GD}, we use Mean Absolute Error (MAE) and Root Mean Squared Error (RMSE) for counting, and Precision, Recall, and F1-score for localization. Localization metric as it verifies precise subclass discrimination, preventing deceptively low counting errors from error cancellation.

\noindent
\textbf{Implementation Details.} 
\label{implementation details}
We train W2-Net for 15 epochs on REC-8K using the AdamW optimizer with a learning rate of $10^{-5}$. Each training batch uses a single image with its corresponding referring expressions. 

More details of the experimental setup are in the supplementary material.

\begin{table}[t]
  \centering

  \resizebox{1.0\linewidth}{!}{
    \begin{tabular}{cc|ccccc}
      \specialrule{0.15em}{0pt}{0pt}
      \rowcolor{mygray}
      W2 Decoder & SSM & MAE~$\downarrow$ & RMSE~$\downarrow$ & Prec~$\uparrow$ & Rec~$\uparrow$ & F1~$\uparrow$ \\
      \specialrule{0.1em}{0pt}{0pt}
       & & 7.21 & 18.69 & 0.63 & 0.69 & 0.66 \\
       & \checkmark & 5.93 & 16.71 & 0.66 & 0.71 & 0.68 \\
      \checkmark & & 3.85 & 11.55 & 0.75 & 0.77 & 0.76 \\
      \rowcolor{lightblue}
      \checkmark & \checkmark &   \textbf{3.55}      &   \textbf{10.39}     &  \textbf{0.76}& \textbf{0.78}   & \textbf{0.77} \\
      \specialrule{0.15em}{0pt}{0pt}
    \end{tabular}
  }
\caption{\textbf{Ablation study on the proposed W2 Decoder and Subclass Separable Matching (SSM).}}
\label{tab: ablation components}
\end{table}

\begin{table}[t]
  \centering
  \resizebox{.7\linewidth}{!}{
    \begin{tabular}{c|cccccc}
      \specialrule{0.15em}{0pt}{0pt}
      \rowcolor{mygray}
      $\lambda_{rep}$ & 0 & 0.1 & 0.2 & 0.5 & 1.0 \\
      \specialrule{0.1em}{0pt}{0pt}
      MAE~$\downarrow$ & 3.85 & 3.67 & \textbf{3.55} & 3.64  & 3.92 \\
      F1~$\uparrow$ & 0.76 & 0.76 & \textbf{0.77} & 0.77 & 0.74 \\
      \specialrule{0.15em}{0pt}{0pt}
    \end{tabular}
  }
\caption{\textbf{Analysis on repulsive weight $\boldsymbol{\lambda_{rep}}$ in SSM.} }
\label{tab: ablation lambda}
\end{table}

\begin{table*}[ht]
\centering

\resizebox{.85\linewidth}{!}{
\begin{tabular}{l|c|c||cc|cc}
\specialrule{0.18em}{0pt}{0pt}
\rowcolor{mygray}   &   &     & \multicolumn{2}{c}{Val Set} & \multicolumn{2}{c}{Test Set}           \\
\cline{4-7}
\rowcolor{mygray} \multirow{-2}{*}{Method}&    \multirow{-2}{*}{Venue} & \multirow{-2}{*}{Prompt} &MAE~$\downarrow$  & RMSE~$\downarrow$     & MAE~$\downarrow$    & RMSE~$\downarrow$  \\ 

\specialrule{0.15em}{0pt}{0pt}

CounTR~\cite{liu2022countr}  &  BMVC22  & Visual       & 13.13      & 49.83      & 11.95 & 91.23                      \\
LOCA~\cite{djukic2023low_LOCA}&  ICCV23    & Visual         & 10.24      & 32.56      & 10.79 & 56.97                      \\
CACViT~\cite{wang2024vision_CACViT}   &  AAAI24  & Visual        & 10.63      & 37.95      & 9.13 & 48.96                      \\
DAVE~\cite{pelhan2024dave_DAVE} &  CVPR24     & Visual       & 8.91       & 28.08      & 8.66  & 32.36                      \\
CountGD~\cite{amini2024countgd_CountGD}   &  NIPS24   & Visual        & \textbf{7.10}      & \textbf{26.08}      & \textbf{5.74} & \textbf{24.09}     \\ 

\specialrule{0.15em}{0pt}{0pt}

Patch-Selection~\cite{xu2023zero_ZSC} &  CVPR23 & Text  & 26.93   & 88.63          & 22.09 & 115.17                     \\
CLIP-Count~\cite{jiang2023clip_CLIP-Count}   &  ACMMM23 & Text    & 18.79      & 61.18      & 17.78 & 106.62 \\
VLCounter~\cite{kang2024vlcounter}&  AAAI24   & Text     & 18.06      & 65.13      & 17.05 & 106.16                     \\
CounTX~\cite{AminiNaieni23_CounTX} &  BMVC23 & Text         & 17.10      & 65.61      & 15.88 & 106.29                     \\
DAVE~\cite{pelhan2024dave_DAVE}  &  CVPR24 & Text          & 15.48      & 52.57      & 14.90 & 103.42                     \\
GroundingREC~\cite{dai2024referring_REC}  &  CVPR24 & Text   & 10.06      & 58.62      & 10.12 & 107.19                     \\
CountGD~\cite{amini2024countgd_CountGD} &  NIPS24  & Text       & 12.14      & 47.51      & 12.98 & 98.35                      \\ 
CAD-GD~\cite{wang2025exploring_CAD-GD}  &  CVPR25   & Text           &  9.30       & 40.96      & 10.35 & 86.88         \\ 


\rowcolor{lightblue} \textbf{W2-Net (Ours)}   & -     &  Text       &   \textbf{8.73}   &  \textbf{37.32}     &  \textbf{9.53} &  \textbf{83.46}                   \\ 
\specialrule{0.18em}{0pt}{0pt}
\end{tabular}}
  \caption{\textbf{Comparison with some state-of-the-art approaches on FSC-147~\cite{ranjan2021learning_FSC}.} 
  The upper and lower parts present the results using visual exemplars and class text as prompts separately.
  }
\label{tab: fsc147}
\end{table*}



\subsection{Comparison with State-of-the-Art}
\label{subsec: comparison with state of the art}
We compare our \textit{W2-Net} with existing methods on the REC-8K benchmark in Tab.\ref{tab: rec-8k val set} and Tab.\ref{tab: rec-8k test set}. \textit{W2-Net} establishes \textbf{new state-of-the-art (SOTA)}, significantly outperforming all prior works on both counting and localization metrics.

With a Swin-T backbone~\cite{liu2021swin}, \textit{W2-Net} reduces the SOTA MAE from 4.58 to 3.55 on the validation set and from 4.59 to 3.59 on the test set, corresponding to a \textbf{relative error reduction of 22.5\% and 18.0\%}, respectively. Concurrently, it boosts the localization F1-score by a significant 7\%.
The performances are more pronounced with a Swin-B backbone, where our method achieves a new record with an MAE of 3.56 and an F1-score of 0.80 on the test set, substantially widening the gap with previous methods.

The substantial gains confirm the success of our approach in tackling the identified annotation challenge. 
By enabling the model to jointly reason about class-representative locations (``what to count'') and attribute-specific visual regions (``where to see''), \textit{W2-Net} learns a highly discriminative representation with more stable supervision provided by our Subclass Separable Matching (SSM). 
This directly translates into reduced subclass confusion and superior localization accuracy, leading to more reliable counting performance and setting a new record for the REC task.
Fig.~\ref{fig:qualitative_results} provides qualitative examples, visually demonstrating that \textit{W2-Net} successfully distinguishes challenging subclasses.

\subsection{Ablation Study}
\label{subsec: ablation study}
All ablation studies are conducted on the REC-8K validation set using the Swin-T backbone.

\noindent
\textbf{Proposed Components.}
We first evaluate the impact of our two main contributions independently. As shown in Tab.~\ref{tab: ablation components}, starting from a baseline MAE of 7.21, integrating our Subclass Separable Matching (SSM) alone reduces the error to 5.93 by stabilizing the training signal. 
Independently, the W2 Decoder yields a more substantial improvement (MAE 3.85, F1 0.76), confirming that explicitly modeling ``where to see'' is crucial for fine-grained localization. 
Combining both components in our full \textit{W2-Net} model yields the best performance, demonstrating their synergistic effect: the W2 Decoder provides superior features, while SSM ensures they are learned through precise label assignment.

\noindent
\textbf{Analysis of Repulsive Weight $\boldsymbol{\lambda_{rep}}$ in Eq.~\ref{eq:matching_cost}.}
We analyze the effect of the repulsive force hyperparameter $\lambda_{rep}$ in SSM.
As presented in Tab.~\ref{tab: ablation lambda},
performance improves as $\lambda_{rep}$ increases from 0 (disabling the repulsive force), peaking at $\lambda_{rep}=0.2$.
A further increase slightly degrades performance, suggesting that an overly strong repulsive force can incorrectly overwhelm other matching cost terms, such as the classification.
We thus set $\lambda_{rep}=0.2$.

\begin{table}[t]
\centering
\resizebox{1.\linewidth}{!}{
\begin{tabular}{l|c||cc}
\specialrule{0.15em}{0pt}{0pt}
\rowcolor{mygray}         &  & \multicolumn{2}{c}{Test Set} \\ 
\cline{3-4}
\rowcolor{mygray}\multirow{-2}{*}{Method}&\multirow{-2}{*}{Prompt} & MAE~$\downarrow$& RMSE~$\downarrow$     \\

\specialrule{0.15em}{0pt}{0pt}

CLIP-Count~\cite{jiang2023clip_CLIP-Count} & Text              & 11.96       & 16.61      \\
CounTX~\cite{AminiNaieni23_CounTX}*     & Text           & 8.13        & 10.87      \\
VLCounter~\cite{kang2024vlcounter}  & Text                   & 6.46        & 8.68       \\
CACViT~\cite{wang2024vision_CACViT}*     & Visual              & 4.91        & 6.49       \\
CountGD~\cite{amini2024countgd_CountGD}    & Text                 & 3.83        & 5.41       \\
CountGD~\cite{amini2024countgd_CountGD}    & Both         & 3.68        & 5.17       \\
CAD-GD~\cite{wang2025exploring_CAD-GD}      & Text                   & 3.29        & 4.56   \\ 
\rowcolor{lightblue} \textbf{W2-Net (Ours)}               & Text     & \textbf{2.89}    &          \textbf{4.13}           \\ 

\specialrule{0.18em}{0pt}{0pt}

\end{tabular}}
\caption{\textbf{Comparison with some state-of-the-art approaches on the CARPK dataset.} * means finetuned.
}
\label{tab: carpk}
\end{table}

\subsection{Results on Zero-shot Counting}
\label{subsec: zero-shot}

\noindent
\textbf{FSC-147~\cite{ranjan2021learning_FSC}.}
To assess the generalizability of our framework, we evaluate \textit{W2-Net} on the FSC-147 benchmark for zero-shot class-agnostic counting. 
As shown in Tab.~\ref{tab: fsc147}, our method achieves strong performance, outperforming previous text-prompted approaches with an MAE of 8.73 and 9.53 on the validation and test sets, respectively. 
It is crucial to note that FSC-147 provides only class-level labels (\textit{e.g.}, ``car''), lacking the fine-grained attributes needed to explicitly guide our w2s query. 
The strong performance, therefore, highlights the inherent robustness of our dual-query design.
Even without explicit attribute supervision, the w2s query acts as a dynamic feature scout, automatically exploring supplementary visual cues that enrich the w2c query's representation. This capability leads to more robust class-level identification.

\noindent
\textbf{CARPK~\cite{hsieh2017drone_carpk}. }
To further test the cross-dataset generalization, we directly evaluate our FSC-147-trained model on the CARPK car counting dataset without any fine-tuning.
The results, presented in Tab.\ref{tab: carpk}, demonstrate that \textit{W2-Net} surpasses existing methods, including those that leverage visual exemplar prompts, demonstrating its excellent transferability.

\section{Conclusion}
\label{sec: conclusion}

In this paper, we identify and address a critical, yet overlooked, challenge in Referring Expression Counting (REC): the misalignment between class-representative annotations and attribute-defining visual regions. We introduce \textit{W2-Net}, a novel framework that resolves this issue by decoupling ``what to count'' and ``where to see'' via its W2 decoder. 
This approach, synergized with our Subclass Separable Matching, significantly outperforms previous REC methods in counting and localization.
Specifically, \textit{W2-Net} achieves a remarkable reduction in relative counting error by $22.5\%$ on the validation set and $18.0\%$ on the test set, while also significantly improving the localization accuracy. 
Beyond these impressive performance gains, we hope this work can shed light on a fundamental annotation challenge in REC.
\renewcommand{\thefigure}{S-\arabic{figure}}
\renewcommand{\thetable}{S-\arabic{table}}

\section{Supplementary Material}

\subsection{Introduction}
\label{supplementary_sec: introduction}
This supplementary material includes following contents:
\begin{itemize}
\item Loss function details.
\item Additional qualitative results on REC-8K of W2 Decoder Mechanism.
\item Experimental setup details.
\item Efficacy analysis.
\item Ablation studies on the repulsive cost formulation in Eq.~9.
\item Limitation and future work.
\end{itemize}

\subsection{Loss Function Details}
\label{supplementary_sec: loss}

The training of our \textit{W2-Net} is optimized using a composite loss function, denoted as $\mathcal{L}$, which is a weighted sum of a classification loss ($\mathcal{L}_{cls}$) and a localization loss ($\mathcal{L}_{loc}$).
All the settings follow prior works, GroundingREC~\cite{dai2024referring_REC} and CAD-GD~\cite{wang2025exploring_CAD-GD}, to ensure a fair comparison. 
The total loss is defined as:
\begin{equation}
    \mathcal{L} = \lambda_{cls}\mathcal{L}_{cls} + \lambda_{loc}\mathcal{L}_{loc}
\end{equation}
The balancing weights are not finetuned and set to $\lambda_{cls} = 5$ and $\lambda_{loc} = 1$. Both loss components are computed on the predictions from the \textit{what-to-count} (w2c) queries, which are assigned to ground-truth instances via our proposed Subclass Separable Matching (SSM).

\paragraph{Classification Loss ($\mathcal{L}_{cls}$)}
We employ the Focal Loss at a fine-grained \textit{query-textword} level. For each of the $K$ w2c queries and $N$ text tokens, the model predicts a cosine similarity. We use the standard Focal Loss with parameters $\gamma=2.0$ and $\alpha=0.25$. The classification loss is the average Focal Loss over all query-token pairs, formulated as:
\begin{equation}
    \mathcal{L}_{cls} = \frac{1}{N_{pos}} \sum_{i=1}^{K} \sum_{j=1}^{N} \text{FL}(s_{ij}, y_{ij})
\end{equation}
where:
\begin{itemize}
    \item $N_{pos}$ is the number of positive queries matched to ground-truth objects.
    \item $s_{ij}$ is the cosine similarity (logit) for the $i$-th query and $j$-th text token.
    \item $y_{ij}$ is the ground-truth label. If query $i$ is a positive match, $y_{ij}=1$ for all its corresponding text tokens $j$; otherwise, $y_{ij}=0$.
\end{itemize}

\paragraph{Localization Loss ($\mathcal{L}_{loc}$)}
The localization loss is applied only to the positive-matched queries. We use the L1 loss to measure the distance between the predicted point location $\hat{p}_i$ and its corresponding ground-truth annotation $p_{\sigma(i)}$:
\begin{equation}
    \mathcal{L}_{loc} = \frac{1}{N_{pos}} \sum_{i \in \sigma} \| \hat{p}_i - p_{\sigma(i)} \|_1
\end{equation}
Here, $\sigma$ represents the set of positive matches determined by SSM, and $\| \cdot \|_1$ is the L1 norm.

\subsection{Additional qualitative results on REC-8K of W2 Decoder Mechanism}
\label{sec: vis_w2_decoder}

\begin{figure*}[t!]
    \centering
    \includegraphics[width=\linewidth]{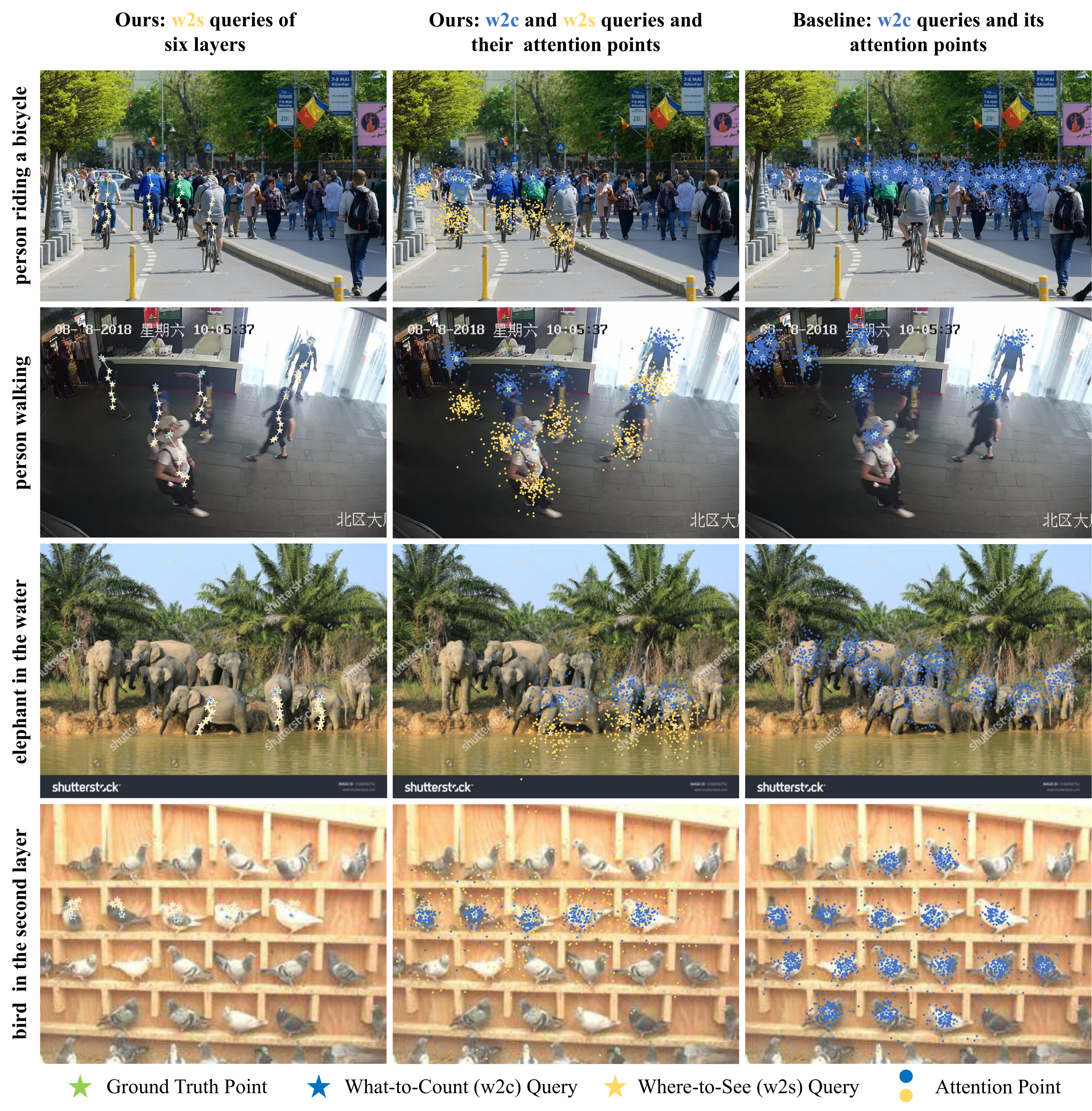} 
    \caption{
        \textbf{Qualitative visualization of the W2-Decoder's mechanism on the REC-8K dataset.}
        Green, blue, and yellow pentastars denote the ground-truth (GT) point, the final predicted point from the what-to-count (w2c) query, and the where-to-see (w2s) query, respectively. Attention points visualize the focus area of each query.
        \textbf{(Left Column):} The trajectory of our w2s query across six decoder layers, demonstrating its progressive convergence towards attribute-relevant regions.
        \textbf{(Middle Column):} Synergy in Our \textit{W2-Net}. The w2c query's attention correctly centers on class-representative areas (e.g., persons' head), while the w2s query's attention seeks out attribute-specific visual cues (e.g., the bicycle for ``person riding a bicycle''). This fusion of attribute-aware features enables precise subclass discrimination.
        \textbf{(Right Column):} The baseline model. Its w2c query, solely supervised by the GT point, focuses narrowly on class features and neglects crucial attribute information, hindering its ability to distinguish between similar subclasses.
        Notably, for the spatial attribute ``bird in the second layer'' (bottom row), our w2s query expands its search area to gather global context, enabling precise counting and localization based on relative position.
    }
    \label{fig:qualitative_w2_decoder}
\end{figure*}

To provide intuitive evidence for the effectiveness of our W2 Decoder, we present a qualitative analysis on some samples from the REC-8K dataset in Fig.~\ref{fig:qualitative_w2_decoder}. These visualizations concretely demonstrate how our proposed dual-query mechanism successfully decouples ``what to count'' (localizing the object) and ``where to see'' (identifying attribute-defining regions), thereby resolving the core annotation-feature misalignment problem identified in our main paper.

As illustrated in Fig.~\ref{fig:qualitative_w2_decoder}, our approach exhibits a clear and effective division of roles between the two query types.

\paragraph{Dynamic Attribute Seeking of the W2S Query.} The leftmost column visualizes the iterative refinement of the w2s query's position through the six decoder layers. In each case, the w2s query (yellow pentastar) starts near the class-representative location but actively moves to seek out the most informative visual region for the specified attribute. For "person riding a bicycle", it converges on the bicycle frame; for ``person walking'', it moves towards the legs; for ``elephant in the water'', it focuses on the water surrounding the elephant, capturing the environmental context. This demonstrates that the w2s query functions as a dedicated ``attribute scout'', as intended.

\paragraph{Synergistic Feature Fusion for Precise Discrimination.} The middle column showcases the different attention point distributions of w2c and w2s queries. The w2c query's attention points remain focused on class-representative regions (e.g., the person's head, the elephant's whole body, the bird's whole body), consistent with the GT supervision. Crucially, the w2s queries attention points lock onto attribute-defining regions. By fusing the features extracted by the w2s query, the w2c query is enriched with the critical information needed to distinguish subclasses. For instance, the bicycle features help differentiate ``riding a bicycle'' from ``standing'', and the leg features help differentiate ``walking'' from ``standing''.

A particularly compelling example is the bottom row (``bird in the second layer''). Here, the attribute is purely spatial and relational. Our w2s query demonstrates remarkable adaptability by not just moving upwards, but also expanding its attention field to aggregate global layout information. This contextual understanding is then passed to the w2c query, enabling it to correctly identify the bird in the specified layer, a task that requires reasoning beyond local object features.

\paragraph{Limitations of the Baseline.} In contrast, the rightmost column reveals the baseline's inherent weakness. Constrained by the GT point annotation on class-representative regions (\textit{e.g.}, ``person head'' and ``bird body''), the baseline's w2c query focuses its attention almost exclusively on that area. While sufficient for class identification (``person'', ``bird''), it fails to gather the necessary attribute-specific information from other regions (e.g., bicycle, legs, global position). This lack of attribute-aware features makes it fundamentally difficult for the baseline to resolve ambiguities between fine-grained subclasses, leading to the performance gap we report in the main paper.

In summary, these qualitative results provide strong visual evidence that our \textit{W2-Net}, through its novel W2 Decoder, effectively learns to see ``where'' the attribute is, which in turn helps it to better count ``what'' the target is.

\subsection{Experimental Setup Details}

\paragraph{Datasets.}
REC-8K~\cite{dai2024referring_REC} is the first REC dataset, whose images are collected from eight existing datasets (FSC-147~\cite{ranjan2021learning_FSC}, JHU Crowd~\cite{sindagi2020jhu}, NWPU~\cite{wang2020nwpu}, VisDrone~\cite{zhu2021detection_VisDrone}, DETRAC~\cite{wen2020ua_DETRAC}, CARPK~\cite{hsieh2017drone_carpk}, Mall~\cite{chen2013cumulative_MALL}, and Crowd Surveillance~\cite{yan2019perspective_CrowdSurveillance})
and photo sharing websites Pixabay and Unsplash.
It contains 8,011 images and 17,122 image-RE pairs, split into training (10555 pairs), validation (3336 pairs), and test (3231 pairs) sets.

\paragraph{Metrics.}
Following prior object counting works, we use Mean Absolute Error (MAE) and Root Mean Squared Error (RMSE) to measure counting performance,  
defined as $\text{MAE} = \frac{1}{N_p} \sum\limits_{i=1}^{N_p} \left| c_i - \hat{c}_i \right|, \quad \text{RMSE} = \sqrt{\frac{1}{N_p} \sum\limits_{i=1}^{N_p} \left(  c_i - \hat{c}_i \right)^2}$, where $N_p$ is the number of image-referring-expression pairs, and $\hat{c}_i$ and $c_i$ denote the predicted count and corresponding ground-truth count for the $i$-th test image-RE pairs. 

For the REC-8K dataset, we also adopt Precision, Recall, and F1 scores as metrics for localization.
A predicted point is considered a True Positive (TP) if its distance from the corresponding ground truth point (GT) is within a threshold; otherwise, it is considered a False Positive (FP). An unmatched ground truth point is considered a False Negative (FN).
The performance of crowd localization is given by:
$\text{Prec} = \frac{\text{TP}}{\text{TP} + \text{FP}}, \quad \text{Rec} = \frac{\text{TP}}{\text{TP} + \text{FN}}, \quad \text{F1} = 2 \cdot \frac{\text{Prec} \cdot \text{Rec}}{\text{Prec} + \text{Rec}}$.

Notably, localization metrics are particularly crucial for REC as models can achieve deceptively low counting errors by balancing false positives and false negatives across visually similar subclasses that share the same base class. 
A high F1-score, therefore, provides a more comprehensive measure of a model's ability to accurately distinguish and localize the correct subclass objects, ensuring good counting performance stems from precise subclass discrimination rather than error cancellation.

\subsubsection{Implementation Details.}
Following GroundingREC~\cite{dai2024referring_REC} and CAD-GD~\cite{wang2025exploring_CAD-GD},
our \textit{W2-Net} is built upon the open-set detector framework of GroundingDINO~\cite{liu2024grounding_gdino}. We use Swin-Transformer~\cite{liu2021swin} (Swin-T and Swin-B variants) as the image backbone and the BERT model as the text backbone.

\paragraph{REC-8K Dataset.}
\begin{itemize}
    \item \textbf{Training:} We train the \textit{W2-Net} model for 15 epochs using the AdamW optimizer with a fixed learning rate of $10^{-5}$ and a weight decay of $10^{-4}$. Each training batch consists of a single image paired with its multiple corresponding referring expressions. We freeze both the visual (Swin-T/Swin-B) and text (BERT) backbones during training to maintain the aligned feature spaces. The loss weights and matching cost weight are set to $\lambda_{cls} = 5$ for the classification loss and $\lambda_{L1} = 1$ for the localization loss, consistent with GroundingREC~\cite{dai2024referring_REC} and CAD-GD~\cite{wang2025exploring_CAD-GD}. 
    For the additionally introduced repulsive force in our SSM, its weight $\lambda_{rep}$ is set to 0.2, which was identified as optimal in our ablation study (see Tab.~4 in the main paper). 
    The number of object queries in the decoder is kept at $K=900$.
    \item \textbf{Inference:} The inference strategy is consistent with GroundingREC~\cite{dai2024referring_REC} and CAD-GD~\cite{wang2025exploring_CAD-GD}, where a w2c query prediction is counted if its score surpasses a threshold of 0.25 for the CLS text token and 0.35 for other text tokens. 
\end{itemize}

\paragraph{FSC-147 and CARPK Datasets (Zero-shot Counting).}
\begin{itemize}
    \item \textbf{Training (on FSC-147):} To evaluate zero-shot performance, the model is trained exclusively on the FSC-147 training set for 30 epochs. 
    Following CAD-GD~\cite{wang2025exploring_CAD-GD}, we use the AdamW optimizer with a batch size of 4 and a learning rate of $1 \times 10^{-5}$, which decays by a factor of 10 at the 15th epoch. The image and text backbones remain frozen. For data augmentation, we follow CAD-GD~\cite{wang2025exploring_CAD-GD} and CountGD~\cite{amini2024countgd_CountGD}, where the image's shorter side is randomly resized to a value in \{480, 512, 544, 576, 608, 640, 672, 704, 736, 768, 800\} pixels while maintaining the aspect ratio. All class labels in the training set are concatenated into a single caption.
    \item \textbf{Inference (on FSC-147):} Images are resized so their short side is 800 pixels and aspect ratio is maintained. 
    We adopt the adaptive cropping strategy from CAD-GD and CountGD: if the initial predicted count exceeds 600, the image is divided into four non-overlapping crops, which are then resized with the short side of 800 pixels. Then these crops are processed individually and the final count is the sum of counts from all crops. The threshold of FSC-147 is set to 0.3, consistent with CAD-GD.
    \item \textbf{Inference (on CARPK):} To test cross-dataset generalization, the model trained on FSC-147 is evaluated directly on CARPK without any fine-tuning. A threshold of 0.15 is used for this dataset, consistent with CAD-GD.
\end{itemize}

\paragraph{Computing Infrastructure.}
All experiments were conducted on a NVIDIA H800 GPU (80GB) running on a Linux server with PyTorch 2.5.1 and CUDA 12.1.

\subsection{Efficacy Analysis}

We provide a comparison of model parameters and computational cost (GFLOPs) with existing REC methods. As detailed in Tab.~\ref{tab: efficacy}, our \textit{W2-Net} introduces a marginal increase in both parameters and GFLOPs. This modest overhead is attributed to our novel W2 Decoder, which incorporates a parallel stream of where-to-see (w2s) queries to enhance attribute perception.

Crucially, this slight increase in computational cost yields a substantial improvement in performance, as evidenced by the significant reduction in counting error and the boost in localization F1-score reported in our main paper. This highlights a highly favorable efficiency-performance trade-off. For practical reference, the full training of \textit{W2-Net} with a Swin-B backbone on the REC-8K dataset takes approximately 7 hours and requires around 23GB of VRAM on a NVIDIA H800 GPU.
Testing on the entire validation and test sets takes approximately 150s and 140s, respectively.

\begin{table}[h]
\centering
\resizebox{1.\linewidth}{!}{
\begin{tabular}{l|c||cc}
\specialrule{0.15em}{0pt}{0pt}
\rowcolor{mygray}     &    &  &  \\ 
\rowcolor{mygray}\multirow{-2}{*}{Method}&\multirow{-2}{*}{Backbone} &\multirow{-2}{*}{Parameters(M)} & \multirow{-2}{*}{GFLOPs}    \\

\specialrule{0.15em}{0pt}{0pt}

GroundingREC  & Swin-T   & 144.1                      & 67.7               \\
CAD-GD        & Swin-T   & 159.6                      & 74.7               \\
W2-Net (Ours) & Swin-T   & 152.0             & 76.6      \\ 
\specialrule{0.15em}{0pt}{0pt}
GroundingREC   & Swin-B   & 204.0                      & 98.6               \\
CAD-GD         & Swin-B   & 219.5                      & 105.5              \\
W2-Net (Ours) & Swin-B   & 211.9             & 107.4     \\ 

\specialrule{0.15em}{0pt}{0pt}

\end{tabular}}
\caption{Comparison of the model size and FLOPs. The FLOPs are obtained using a $3 \times 384 \times 384$ image with the referring expression ``person walking'' as input.
}
\label{tab: efficacy}
\end{table}

\begin{table*}[th]
\centering
\resizebox{.9\linewidth}{!}{
\begin{tabular}{l|c||ccccc}
\specialrule{0.15em}{0pt}{0pt}
\rowcolor{mygray}      &  & & & & & \\ 
\rowcolor{mygray}
\multirow{-2}{*}{Repulsive Cost Formulation} & \multirow{-2}{*}{Expression $C_{rep}$} & \multirow{-2}{*}{MAE $\downarrow$} & \multirow{-2}{*}{RMSE $\downarrow$} & \multirow{-2}{*}{Prec $\uparrow$} & \multirow{-2}{*}{Rec $\uparrow$} & \multirow{-2}{*}{F1 $\uparrow$} \\
\specialrule{0.15em}{0pt}{0pt}
No Repulsion (Baseline) & N/A & 7.21 & 18.69 & 0.63 & 0.69 & 0.66 \\
\hline
Inverse Negative Distance & $1 / (d_{neg} + \epsilon)$ & 7.05 & 18.45 & 0.65 & 0.67 & 0.66 \\
Normalized Positive Distance & $d_{pos} / (d_{pos} + d_{neg})$ & 6.88 & 18.12 & 0.65 & 0.70 & 0.67 \\
Hinge on Ratio & $\max(0, 1 - R)$ & 6.41 & 17.54 & 0.66 & 0.69 & 0.67 \\
\textbf{Proposed (Exponential Ratio)} & $\boldsymbol{\exp(-R)}$ & \textbf{5.93} & \textbf{16.71} & \textbf{0.66} & \textbf{0.71} & \textbf{0.68} \\
\specialrule{0.15em}{0pt}{0pt}
\end{tabular}}
\caption{Ablation study on the formulation of the repulsive cost in SSM. All experiments are conducted on the REC-8K validation set with Swin-T backbone, without the W2 Decoder, to isolate the effect of the repulsive cost. Our proposed exponential formulation demonstrates superior performance over other intuitive alternatives. Here $d_{pos} = \|\hat{p}_i - p_j\|_2$, $d_{neg} = \min_{p_k \in Y_{neg}} \|\hat{p}_i - p_k\|_2$, and $R = d_{neg}/d_{pos}$.}
\label{tab:repulsive_cost_formulation}
\end{table*}

\subsection{Ablation Study on the Repulsive Cost Formulation}
\label{supplementary_sec:repulsive_cost_ablation}

In the main paper, our Subclass Separable Matching (SSM) introduces a repulsive cost term, $C_{rep}$, to mitigate matching ambiguity. The formulation we proposed is based on an exponential function of the ambiguity ratio $R(\hat{p}_i, p_j)$, as defined in Eq.~(8) and Eq.~(9):
\begin{equation}
    \mathcal{R}(\hat{p}_i, p_j) = \frac{\min_{p_k \in Y_{neg}} \|\hat{p}_i - p_k\|_2}{\|\hat{p}_i - p_j\|_2}~.
    \label{eq:ambiguity_ratio}
\end{equation}
\begin{equation}
    \mathcal{C}_{rep}(\hat{p}_i, p_j) = \exp(-\mathcal{R}(\hat{p}_i, p_j))~.
    \label{eq:repulsive_cost}
\end{equation}
This formulation provides a smooth but strong penalty when a prediction is closer to a non-target object than its potential ground-truth match (i.e., when $R < 1$). To validate this design choice, we conduct an ablation study to compare it against other plausible formulations for the repulsive cost. All experiments are run on the REC-8K validation set with the Swin-T backbone, and the repulsive weight $\lambda_{rep}$ is kept at 0.2 for all variants.

We compare the following formulations:
\begin{itemize}
    \item \textbf{No Repulsion:} The baseline where $C_{rep}$ is omitted (i.e., $\lambda_{rep}=0$). This is equivalent to the Baseline setting without W2 Decoder and SSM in Tab.~3 of the main paper.
    \item \textbf{Inverse Negative Distance:} A simpler formulation that only penalizes proximity to any non-target object, defined as $C_{rep} = 1 / (d_{neg} + \epsilon)$, where $d_{neg} = \min_{p_k \in Y_{neg}} \|\hat{p}_i - p_k\|_2$ and $\epsilon$ is a small constant.
    \item \textbf{Normalized Positive Distance:} This formulation calculates the cost as the ratio of the distance to the positive ground truth against the sum of distances to both the positive and the nearest negative ground truth. It is defined as $C_{rep} = d_{pos} / (d_{pos} + d_{neg})$, where $d_{pos} = \|\hat{p}_i - p_j\|_2$. This provides a naturally normalized cost in the range $[0, 1]$.
    \item \textbf{Hinge on Ratio:} A linear penalty based on the ambiguity ratio $R$, defined as $C_{rep} = \max(0, 1 - R)$. This provides a penalty only when the prediction is closer to a distractor ($R < 1$).
    \item \textbf{Proposed (Exponential Ratio):} The formulation used in our final \textit{W2-Net} model, $C_{rep} = \exp(-R)$.
\end{itemize}

The results are presented in Tab.~\ref{tab:repulsive_cost_formulation}. Among these, our proposed exponential formulation achieves the best performance across all metrics. Its superiority lies in how it handles ambiguity compared to the alternatives. The \textbf{Hinge on Ratio} method, for instance, is indifferent once a match is "good enough" (i.e., when $R>1$, the penalty and its effect become zero), providing no incentive to further push the prediction away from potential distractors. The Normalized Positive Distance ($d_{pos}/(d_{pos}+d_{neg})$), while providing a continuous signal, suffers from a different limitation: its repulsive force increases in a near-linear fashion as a prediction moves closer to a negative instance. It fails to amplify the penalty for the most critical cases where a query is significantly closer to a non-target object than to its correct target.

In contrast, our proposed $\boldsymbol{\exp(-R)}$ formulation provides a sharp, non-linear penalty that is most pronounced precisely in these critical regions. By using an exponential function, the cost increases dramatically as the ambiguity ratio $R$ drops below 1, effectively delivering a decisive ``push'' to resolve hard ambiguous assignments. This ability to dynamically scale the penalty based on the severity of ambiguity ensures a cleaner training signal.

\subsection{Limitation and Future Work}
\label{supplementary_sec: limitation}
A primary limitation of our method
is the fixed number of object queries ($K=900$),
which constrains its direct application to extremely dense scenes where the number of instances exceeds this limit.
While a standard tiling or cropping strategy can serve as a workaround, it introduces more computational overhead.
Therefore, a potential direction for future work lies in developing an adaptive query generation mechanism, especially for extremely dense scenes.

\bibliography{aaai2026}

\begin{thebibliography}{51}
\providecommand{\natexlab}[1]{#1}

\bibitem[{Amini-Naieni et~al.(2023)Amini-Naieni, Amini-Naieni, Han, and Zisserman}]{AminiNaieni23_CounTX}
Amini-Naieni, N.; Amini-Naieni, K.; Han, T.; and Zisserman, A. 2023.
\newblock Open-world Text-specified Object Counting.
\newblock In \emph{BMVC}.

\bibitem[{Amini-Naieni et~al.(2024)Amini-Naieni, Han, Zisserman, and .}]{amini2024countgd_CountGD}
Amini-Naieni, N.; Han, T.; Zisserman, A.; and . 2024.
\newblock CountGD: Multi-modal open-world counting.
\newblock In \emph{NIPS}, volume~37, 48810--48837.

\bibitem[{Chang et~al.(2022)Chang, Yujie, Andrew, and Weidi}]{liu2022countr}
Chang, L.; Yujie, Z.; Andrew, Z.; and Weidi, X. 2022.
\newblock CounTR: Transformer-based Generalised Visual Counting.
\newblock In \emph{BMVC}.

\bibitem[{Chen et~al.(2013)Chen, Gong, Xiang, and Change~Loy}]{chen2013cumulative_MALL}
Chen, K.; Gong, S.; Xiang, T.; and Change~Loy, C. 2013.
\newblock Cumulative attribute space for age and crowd density estimation.
\newblock In \emph{CVPR}, 2467--2474.

\bibitem[{Chen et~al.(2025)Chen, Huo, Jiang, Hu, and Chen}]{chen2025single}
Chen, X.; Huo, S.; Jiang, B.; Hu, H.; and Chen, X. 2025.
\newblock Single Domain Generalization for Few-Shot Counting via Universal Representation Matching.
\newblock In \emph{CVPR}, 4639--4649.

\bibitem[{Chen et~al.(2021)Chen, Liang, Bai, Xu, and Yang}]{chen2021cell_yajie}
Chen, Y.; Liang, D.; Bai, X.; Xu, Y.; and Yang, X. 2021.
\newblock Cell localization and counting using direction field map.
\newblock \emph{IEEE JBHI}, 26(1): 359--368.

\bibitem[{Dai, Liu, and Cheung(2024)}]{dai2024referring_REC}
Dai, S.; Liu, J.; and Cheung, N.-M. 2024.
\newblock Referring expression counting.
\newblock In \emph{CVPR}, 16985--16995.

\bibitem[{Dumery et~al.(2025)Dumery, Ett{\'e}, Fan, Li, Xu, Le, and Fua}]{dumery2024counting}
Dumery, C.; Ett{\'e}, N.; Fan, A.; Li, R.; Xu, J.; Le, H.; and Fua, P. 2025.
\newblock Counting Stacked Objects.
\newblock In \emph{CVPR}.

\bibitem[{Guo, Gao, and Yuan(2025)}]{guo2025enhancing_pruning}
Guo, H.; Gao, J.; and Yuan, Y. 2025.
\newblock Enhancing Low-Rank Adaptation with Recoverability-Based Reinforcement Pruning for Object Counting.
\newblock In \emph{AAAI}, volume~39, 3238--3246.

\bibitem[{Han et~al.(2023)Han, Bai, Liu, and Ouyang}]{han2023steerer_STEERER}
Han, T.; Bai, L.; Liu, L.; and Ouyang, W. 2023.
\newblock {STEERER}: {R}esolving Scale Variations for Counting and Localization via Selective Inheritance Learning.
\newblock In \emph{ICCV}, 21848--21859.

\bibitem[{He et~al.(2024)He, Dai, Trigoni, Chen, and Markham}]{he2024soundcount}
He, Y.; Dai, Z.; Trigoni, N.; Chen, L.; and Markham, A. 2024.
\newblock SoundCount: sound counting from raw audio with dyadic decomposition neural network.
\newblock In \emph{AAAI}, volume~38, 12421--12429.

\bibitem[{Hsieh, Lin, and Hsu(2017)}]{hsieh2017drone_carpk}
Hsieh, M.-R.; Lin, Y.-L.; and Hsu, W.~H. 2017.
\newblock Drone-based object counting by spatially regularized regional proposal network.
\newblock In \emph{ICCV}, 4145--4153.

\bibitem[{Huang et~al.(2024)Huang, Nguyen, Nguyen, Pham, and Hoai}]{huang2024count_action}
Huang, Y.; Nguyen, D.~D.; Nguyen, L.; Pham, C.; and Hoai, M. 2024.
\newblock Count what you want: exemplar identification and few-shot counting of human actions in the wild.
\newblock In \emph{AAAI}, volume~38, 10057--10065.

\bibitem[{Idrees et~al.(2018)Idrees, Tayyab, Athrey, Zhang, Al-Maadeed, Rajpoot, and Shah}]{idrees2018composition_UCF-QNRF}
Idrees, H.; Tayyab, M.; Athrey, K.; Zhang, D.; Al-Maadeed, S.; Rajpoot, N.; and Shah, M. 2018.
\newblock Composition loss for counting, density map estimation and localization in dense crowds.
\newblock In \emph{ECCV}, 532--546.

\bibitem[{Jiang, Liu, and Chen(2023)}]{jiang2023clip_CLIP-Count}
Jiang, R.; Liu, L.; and Chen, C. 2023.
\newblock Clip-count: Towards text-guided zero-shot object counting.
\newblock In \emph{ACMMM}, 4535--4545.

\bibitem[{Kang et~al.(2024)Kang, Moon, Kim, and Heo}]{kang2024vlcounter}
Kang, S.; Moon, W.; Kim, E.; and Heo, J.-P. 2024.
\newblock Vlcounter: Text-aware visual representation for zero-shot object counting.
\newblock In \emph{AAAI}, volume~38, 2714--2722.

\bibitem[{Kuhn(1955)}]{kuhn1955hungarian}
Kuhn, H.~W. 1955.
\newblock The Hungarian method for the assignment problem.
\newblock \emph{Naval Research Logistics Quarterly}, 2(1-2): 83--97.

\bibitem[{Li et~al.(2014)Li, Chang, Wang, Ni, Hong, and Yan}]{urban_planning}
Li, T.; Chang, H.; Wang, M.; Ni, B.; Hong, R.; and Yan, S. 2014.
\newblock Crowded scene analysis: {A} survey.
\newblock \emph{IEEE TCSVT}, 25(3): 367--386.

\bibitem[{Liang, Xu, and Bai(2022)}]{liang2022end_CLTR}
Liang, D.; Xu, W.; and Bai, X. 2022.
\newblock An end-to-end transformer model for crowd localization.
\newblock In \emph{ECCV}, 38--54.

\bibitem[{Lin et~al.(2024)Lin, Ma, Hong, Shangguan, and Meng}]{lin2024gramformer}
Lin, H.; Ma, Z.; Hong, X.; Shangguan, Q.; and Meng, D. 2024.
\newblock Gramformer: Learning Crowd Counting via Graph-Modulated Transformer.
\newblock \emph{AAAI}.

\bibitem[{Lin and Chan(2024)}]{lin2024fixed}
Lin, W.; and Chan, A.~B. 2024.
\newblock A fixed-point approach to unified prompt-based counting.
\newblock In \emph{AAAI}, volume~38, 3468--3476.

\bibitem[{Lin, Zhao, and Chan(2025)}]{lin2025point}
Lin, W.; Zhao, C.; and Chan, A.~B. 2025.
\newblock Point-to-Region Loss for Semi-Supervised Point-Based Crowd Counting.
\newblock In \emph{CVPR}, 29363--29373.

\bibitem[{Liu et~al.(2023)Liu, Lu, Cao, and Liu}]{liu2023point_PET}
Liu, C.; Lu, H.; Cao, Z.; and Liu, T. 2023.
\newblock Point-{Q}uery Quadtree for Crowd Counting, Localization, and More.
\newblock In \emph{ICCV}, 1676--1685.

\bibitem[{Liu et~al.(2024)Liu, Zeng, Ren, Li, Zhang, Yang, Jiang, Li, Yang, Su et~al.}]{liu2024grounding_gdino}
Liu, S.; Zeng, Z.; Ren, T.; Li, F.; Zhang, H.; Yang, J.; Jiang, Q.; Li, C.; Yang, J.; Su, H.; et~al. 2024.
\newblock Grounding dino: Marrying dino with grounded pre-training for open-set object detection.
\newblock In \emph{ECCV}, 38--55. Springer.

\bibitem[{Liu et~al.(2021)Liu, Lin, Cao, Hu, Wei, Zhang, Lin, and Guo}]{liu2021swin}
Liu, Z.; Lin, Y.; Cao, Y.; Hu, H.; Wei, Y.; Zhang, Z.; Lin, S.; and Guo, B. 2021.
\newblock Swin transformer: Hierarchical vision transformer using shifted windows.
\newblock In \emph{ICCV}, 10012--10022.

\bibitem[{Ma et~al.(2019)Ma, Wei, Hong, and Gong}]{ma2019bayesian_BL}
Ma, Z.; Wei, X.; Hong, X.; and Gong, Y. 2019.
\newblock Bayesian loss for crowd count estimation with point supervision.
\newblock In \emph{ICCV}, 6142--6151.

\bibitem[{Mondal et~al.(2025)Mondal, Nag, Zhu, and Dutta}]{mondal2025omnicount}
Mondal, A.; Nag, S.; Zhu, X.; and Dutta, A. 2025.
\newblock Omnicount: Multi-label object counting with semantic-geometric priors.
\newblock In \emph{AAAI}, volume~39, 19537--19545.

\bibitem[{Pelhan et~al.(2024)Pelhan, Zavrtanik, Kristan et~al.}]{pelhan2024dave_DAVE}
Pelhan, J.; Zavrtanik, V.; Kristan, M.; et~al. 2024.
\newblock DAVE-A Detect-and-Verify Paradigm for Low-Shot Counting.
\newblock In \emph{CVPR}, 23293--23302.

\bibitem[{Perez, Maji, and Sheldon(2024)}]{perez2024discount_buildingcount}
Perez, G.; Maji, S.; and Sheldon, D. 2024.
\newblock DISCount: counting in large image collections with detector-based importance sampling.
\newblock In \emph{AAAI}, volume~38, 22294--22302.

\bibitem[{Pothiraj et~al.(2025)Pothiraj, Stengel-Eskin, Cho, and Bansal}]{pothiraj2025capture}
Pothiraj, A.; Stengel-Eskin, E.; Cho, J.; and Bansal, M. 2025.
\newblock Capture: Evaluating spatial reasoning in vision language models via occluded object counting.
\newblock In \emph{CVPR}.

\bibitem[{Qian et~al.(2025)Qian, Guo, Deng, Lei, Zhao, Lau, Hong, and Pound}]{qian2025t2icount}
Qian, Y.; Guo, Z.; Deng, B.; Lei, C.~T.; Zhao, S.; Lau, C.~P.; Hong, X.; and Pound, M.~P. 2025.
\newblock T2icount: Enhancing cross-modal understanding for zero-shot counting.
\newblock In \emph{CVPR}, 25336--25345.

\bibitem[{Ranjan et~al.(2021)Ranjan, Sharma, Nguyen, and Hoai}]{ranjan2021learning_FSC}
Ranjan, V.; Sharma, U.; Nguyen, T.; and Hoai, M. 2021.
\newblock Learning to count everything.
\newblock In \emph{CVPR}, 3394--3403.

\bibitem[{Shi et~al.(2025)Shi, Zhang, Yue, Luo, Zhao, and Li}]{shi2025text}
Shi, M.; Zhang, X.; Yue, Z.; Luo, Y.; Zhao, C.; and Li, L. 2025.
\newblock Text-promptable Object Counting via Quantity Awareness Enhancement.
\newblock \emph{arXiv preprint arXiv:2507.06679}.

\bibitem[{Sindagi, Yasarla, and Patel(2020)}]{sindagi2020jhu}
Sindagi, V.~A.; Yasarla, R.; and Patel, V.~M. 2020.
\newblock Jhu-crowd++: {L}arge-scale crowd counting dataset and a benchmark method.
\newblock \emph{IEEE TPAMI}, 44(5): 2594--2609.

\bibitem[{Song et~al.(2021)Song, Wang, Jiang, Wang, Tai, Wang, Li, Huang, and Wu}]{song2021rethinking_P2P}
Song, Q.; Wang, C.; Jiang, Z.; Wang, Y.; Tai, Y.; Wang, C.; Li, J.; Huang, F.; and Wu, Y. 2021.
\newblock Rethinking counting and localization in crowds: {A} purely point-based framework.
\newblock In \emph{ICCV}, 3365--3374.

\bibitem[{Sun et~al.(2023)Sun, An, Liu, Liu, Sakaridis, Fan, and Van~Gool}]{sun2023indiscernible_underwater}
Sun, G.; An, Z.; Liu, Y.; Liu, C.; Sakaridis, C.; Fan, D.-P.; and Van~Gool, L. 2023.
\newblock Indiscernible Object Counting in Underwater Scenes.
\newblock In \emph{CVPR}, 13791--13801.

\bibitem[{Triaridis et~al.(2025)Triaridis, Kaliosis, Nguyen, Xu, Le, and Samaras}]{triaridis2025improving_REC_newwork}
Triaridis, K.; Kaliosis, P.; Nguyen, E.-R.; Xu, J.; Le, H.; and Samaras, D. 2025.
\newblock Improving Contrastive Learning for Referring Expression Counting.
\newblock \emph{arXiv preprint arXiv:2505.22850}.

\bibitem[{{\DJ}uki{\'c} et~al.(2023){\DJ}uki{\'c}, Luke{\v{z}}i{\v{c}}, Zavrtanik, and Kristan}]{djukic2023low_LOCA}
{\DJ}uki{\'c}, N.; Luke{\v{z}}i{\v{c}}, A.; Zavrtanik, V.; and Kristan, M. 2023.
\newblock A low-shot object counting network with iterative prototype adaptation.
\newblock In \emph{ICCV}, 18872--18881.

\bibitem[{Wan, Wu, and Chan(2023)}]{wan2023modeling_NoiseCC_Tpami}
Wan, J.; Wu, Q.; and Chan, A.~B. 2023.
\newblock Modeling Noisy Annotations for Point-Wise Supervision.
\newblock \emph{IEEE TPAMI}, 45(12): 15065--15080.

\bibitem[{Wang et~al.(2025{\natexlab{a}})Wang, Zhou, Dai, Buys, and Gong}]{wang2024enhancing}
Wang, M.; Zhou, J.; Dai, Y.; Buys, E.; and Gong, M. 2025{\natexlab{a}}.
\newblock Enhancing zero-shot counting via language-guided exemplar learning.
\newblock In \emph{CVPR}.

\bibitem[{Wang et~al.(2020)Wang, Gao, Lin, and Li}]{wang2020nwpu}
Wang, Q.; Gao, J.; Lin, W.; and Li, X. 2020.
\newblock NWPU-crowd: A large-scale benchmark for crowd counting and localization.
\newblock \emph{IEEE TPAMI}, 43(6): 2141--2149.

\bibitem[{Wang et~al.(2025{\natexlab{b}})Wang, Pan, Peng, Cheng, Xiao, Jiang, and Cao}]{wang2025exploring_CAD-GD}
Wang, Z.; Pan, Z.; Peng, Z.; Cheng, J.; Xiao, L.; Jiang, W.; and Cao, Z. 2025{\natexlab{b}}.
\newblock Exploring Contextual Attribute Density in Referring Expression Counting.
\newblock In \emph{CVPR}.

\bibitem[{Wang et~al.(2024)Wang, Xiao, Cao, and Lu}]{wang2024vision_CACViT}
Wang, Z.; Xiao, L.; Cao, Z.; and Lu, H. 2024.
\newblock Vision transformer off-the-shelf: a surprising baseline for few-shot class-agnostic counting.
\newblock In \emph{AAAI}, volume~38, 5832--5840.

\bibitem[{Wen et~al.(2020)Wen, Du, Cai, Lei, Chang, Qi, Lim, Yang, and Lyu}]{wen2020ua_DETRAC}
Wen, L.; Du, D.; Cai, Z.; Lei, Z.; Chang, M.-C.; Qi, H.; Lim, J.; Yang, M.-H.; and Lyu, S. 2020.
\newblock UA-DETRAC: A new benchmark and protocol for multi-object detection and tracking.
\newblock \emph{Computer Vision and Image Understanding}, 193: 102907.

\bibitem[{Xu et~al.(2023)Xu, Le, Nguyen, Ranjan, and Samaras}]{xu2023zero_ZSC}
Xu, J.; Le, H.; Nguyen, V.; Ranjan, V.; and Samaras, D. 2023.
\newblock Zero-shot object counting.
\newblock In \emph{CVPR}, 15548--15557.

\bibitem[{Yan et~al.(2019)Yan, Yuan, Zuo, Tan, Wang, Wen, and Ding}]{yan2019perspective_CrowdSurveillance}
Yan, Z.; Yuan, Y.; Zuo, W.; Tan, X.; Wang, Y.; Wen, S.; and Ding, E. 2019.
\newblock Perspective-guided convolution networks for crowd counting.
\newblock In \emph{ICCV}, 952--961.

\bibitem[{Yang et~al.(2025{\natexlab{a}})Yang, Geng, Peng, and Xu}]{yang2025pbecount}
Yang, C.; Geng, T.; Peng, J.; and Xu, C. 2025{\natexlab{a}}.
\newblock PBECount: Prompt-Before-Extract Paradigm for Class-Agnostic Counting.
\newblock In \emph{AAAI}, volume~39, 9139--9147.

\bibitem[{Yang et~al.(2025{\natexlab{b}})Yang, Li, Zhang, Qi, and Shi}]{yang2025taste}
Yang, M.; Li, Z.; Zhang, J.; Qi, L.; and Shi, Y. 2025{\natexlab{b}}.
\newblock Taste More, Taste Better: Diverse Data and Strong Model Boost Semi-Supervised Crowd Counting.
\newblock In \emph{CVPR}, 24440--24451.

\bibitem[{Zhang et~al.(2016)Zhang, Zhou, Chen, Gao, and Ma}]{zhang2016single_SH_MCNN}
Zhang, Y.; Zhou, D.; Chen, S.; Gao, S.; and Ma, Y. 2016.
\newblock Single-image crowd counting via multi-column convolutional neural network.
\newblock In \emph{CVPR}, 589--597.

\bibitem[{Zhu et~al.(2021)Zhu, Wen, Du, Bian, Fan, Hu, and Ling}]{zhu2021detection_VisDrone}
Zhu, P.; Wen, L.; Du, D.; Bian, X.; Fan, H.; Hu, Q.; and Ling, H. 2021.
\newblock Detection and tracking meet drones challenge.
\newblock \emph{IEEE TPAMI}, 44(11): 7380--7399.

\bibitem[{Zou et~al.(2024)Zou, Xiao, Zhou, Sun, Du, and Xu}]{zou2024_SAE}
Zou, Y.; Xiao, X.; Zhou, P.; Sun, Z.; Du, B.; and Xu, Y. 2024.
\newblock Shifted autoencoders for point annotation restoration in object counting.
\newblock In \emph{ECCV}, 113--130. Springer.

\end{thebibliography}

\end{document}